\documentclass[runningheads]{llncs}
\usepackage{mathrsfs}
\usepackage[T1]{fontenc}
\usepackage{amsmath}
\usepackage{amsfonts}
\usepackage{booktabs}
\usepackage{multirow}
\usepackage{graphicx}
\usepackage{hyperref}
\usepackage{bm}
\usepackage{tikz}
\usepackage{xcolor}

\definecolor{checkgreen}{HTML}{34b233}
\newcommand{\tablecheck}{}%
\DeclareRobustCommand{\tablecheck}{%
  \tikz\fill[scale=0.24, color=checkgreen]
  (0,.35) -- (.25,0) -- (1,.9) -- (.25,.20) -- cycle;%
}
\newcommand{\nocheck}{}%
\DeclareRobustCommand{\nocheck}{%
\tikz[scale=0.13,color=red] {
    \draw[line width=0.65,line cap=round] (0,0) to [bend left=6] (1,1);
    \draw[line width=0.65,line cap=round] (0.2,0.95) to [bend right=3] (0.8,0.05);
}
}
\newcommand{\expnumber}[2]{{#1}\mathrm{e}{#2}}
\usepackage{graphicx,verbatim}
\usepackage{siunitx}
\begin{document}
\title{Strategies for Robust Deep Learning Based Deformable Registration}

\titlerunning{Robust Deep Learning Based Deformable Registration}
%
\begin{comment}

\end{comment}

\author{Joel Honkamaa\inst{1}\orcidID{0000-0003-1532-9848} \and Pekka Marttinen\inst{1}\orcidID{0000-0001-7078-7927
}}
%index{Honkamaa, Joel}
%index{Marttinen, Pekka}
\authorrunning{Honkamaa and Marttinen}
\institute{Aalto University, Finland}

\maketitle              % typeset the header of the contribution
\begin{abstract}
Deep learning based deformable registration methods have become popular in recent years. However, their ability to generalize beyond training data distribution can be poor, significantly hindering their usability. LUMIR brain registration challenge for Learn2Reg 2025 aims to advance the field by evaluating the performance of the registration on contrasts and modalities different from those included in the training set. Here we describe our submission to the challenge, which proposes a very simple idea for significantly improving robustness by transforming the images into MIND feature space before feeding them into the model. In addition, a special ensembling strategy is proposed that shows a small but consistent improvement.


\end{abstract}

\section{Introduction}
Deep learning based medical image registration methods have emerged as a strong alternative for classical iterative methods, but their usability has been brought to question due to their potentially poor performance on data outside the training distribution \cite{jena2024deep}. However, the best methods submitted for the LUMIR MRI brain registration challenge organized as part of Learn2Reg 2024 showed strong robustness to domain shifts, failing only on out-of-distribution contrasts. LUMIR challenge for Learn2reg 2025 aims to advance the field particularly in this regard: For training, one is required to use the provided T1-weighted brain MRI images but the evaluation is performed on new contrasts or even modalities. This paper is an algorithm description of our submission to the challenge.

As our main contribution, we propose to transform the images using the MIND \cite{heinrich2012mind} transformation before feeding them into the model, while still using intra-modality similarity loss (normalized cross-correlation) as the training signal. While the MIND features contain less information than the original images, the transformation unifies the representation between different modalities, and the performance on in-domain images remains similar. Earlier MIND features or its variants have been used for evaluating multi-modal similarity (including in deep learning \cite{guo2019multi,hering2022learn2reg,chen2022transmorph}) but to our knowledge they have not been used as an input transformation in deep learning before. In addition, we propose a special ensembling strategy which still retains the diffeomorphic properties of the used backbone model.

\begin{figure}[t]
\centering
\includegraphics[width=1.0\textwidth]{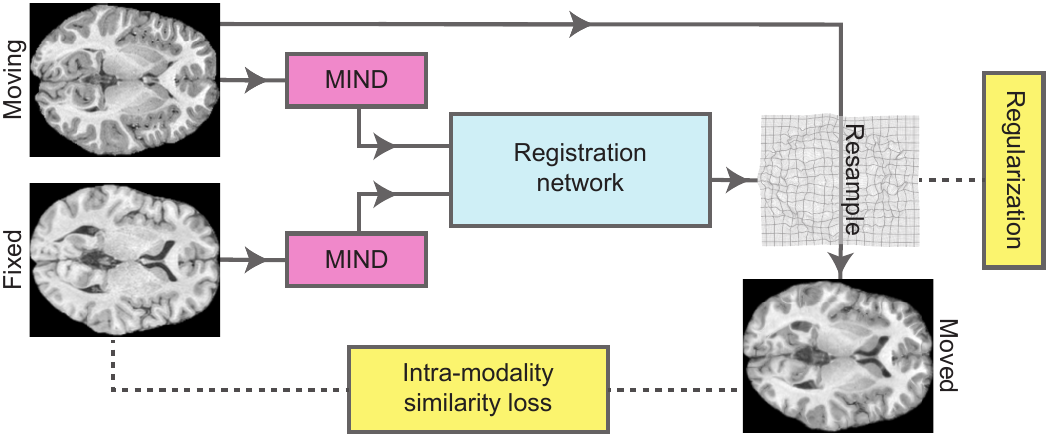}
\caption{Overview of the proposed main idea. The input images go through the MIND transformation before being fed to the registration network. As a result, the network learns to do multi-modal registration even though it is trained with an intra-modality similarity loss. The similarity loss is normalized cross-correlation. Also note that in practice the registration network predicts the deformation in both directions, and the losses are also computed for both directions.}
\end{figure}

\section{Background}\label{sec:background}

MIND (Modality Independent Neighborhood Descriptor) \cite{heinrich2012mind} is a well-established and simple method for measuring multi-modal similarity. The method works by computing MIND features of both images and then taking some simple distance measure such as the mean absolute error or mean squared error between the resulting volumes. MIND encodes how similar a voxel's neighborhood is to its surrounding neighborhoods, not the absolute intensity values.

Given an offset $r$, the formula for computing a single MIND feature at location $x$ can be written as
\begin{equation}\label{eq:MIND}
    \operatorname{MIND}(x, r) = \exp\left(-\frac{D(I, x, x + r)}{V(I, x)}\right)
\end{equation}
where $D(I, x, x + r)$ is the Gaussian-weighted sum of the squared differences between the patches around $x$ and $x + r$
\begin{equation}\label{eq:MIND_MSE}
D(I, x, y) := \sum_{p\in P} \exp(-\frac{p^2}{\sigma^2}) (I(x + p) - I(y + p))^2
\end{equation}
with $P$ being large enough lattice around origin to incorporate most of the Gaussian, and $V(I, x)$ is local variance of $I$ estimated as the mean of $D$ in the six-neighborhood around $x$, giving $V(I, x) := \frac{1}{6} \sum_{n\in \mathscr{N}} D(I, x, x + n)$. Here, $I$ refers to the image, and $\mathscr{N}$ is the six-neighborhood around origin. To compute MIND features, Eq. \ref{eq:MIND} is evaluated for each voxel and multiple offsets, and each voxel is associated with a feature vector consisting of those values. The six-neighborhood set is often used for the offsets as well, resulting in a six-dimensional feature vector.

\section{Methods}

\subsection{Backbone}

As a backbone architecture we use our work SITReg \cite{honkamaa2023sitreg} which was used by the winning submission for Learn2Reg 2024. The architecture is by construction symmetric, inverse consistent, and produces diffeomorphic deformations. The overall architecture starts by extracting multiresolution features from both images independently using ResNet-style convolutional neural network. The architecture then recursively updates the deformation at each resolution starting from the lowest resolution. At each resolution stage, the features of that resolution are transformed by the deformation learned up to that point and are then used to predict a deformation update in symmetric manner. The update deformations are generated using constrained B-spline control points to ensure diffeomorphic predictions. See the paper for more details.

\subsection{Input transformation}

The challenge requires the method to work on images of different contrast or modality from the training images. In general, the behavior of machine learning algorithms on inputs outside the training distribution is hard to predict, and for that reason we take the approach of trying to transform the images to some representation which contains less information than the original representation but is similar across contrasts and modalities. Preferably, mainly the structural information would be preserved. The MIND transformation described in Section \ref{sec:background} is a well-established and simple transformation that unifies different modalities. Note that unlike in the usual use case, we do not use the MIND transformation in computing the similarity loss which is instead computed with the original images using intra-modality loss. Since MIND features unify representation across modalities, the symmetric nature of the backbone architecture is still meaningful even for multi-modal registration. We use $\sigma=0.5$ for the MIND transformation (Eq. \ref{eq:MIND_MSE}) which performed the best in the original paper\cite{heinrich2012mind}.

\subsection{Ensembling}

We train an ensemble of 5 models with different data generation seeds. For the final prediction we average the predicted update deformations at each registration stage of the SITReg multiresolution architecture. We perform averaging in the B-spline weight space to preserve the diffeomorphic properties of the architecture.

\subsection{Further details}

We also use augmentations to help with generalization. We randomly apply Gaussian noise, Gaussian blur, sign inversion, and gamma correction to the input images.

We use normalized cross-correlation as a similarity metric. Due to the MIND input transformation, the network still learns to register images of different modality. For computing the similarity loss, we always use the original non-augmented images, and mask the background out. We regularize the predicted deformations with diffusion regularization (L$^2$ norm on displacements). While the original SITReg paper applied the losses only after the final stage of its multi-resolution architecture, we apply the loss also on intermediate stages to ensure consistent behavior across the trained model ensemble. However, we use very low loss weight of $\frac{1}{100}$ for the earlier stages.

While the SITReg backbone produces nearly perfectly diffeomorphic deformations, due to resampling errors tiny folding errors can still occur. To ensure a very high competency, we add non-diffeomorphic volume (NDV) \cite{liu2024finite} as an additional loss term for the final epochs. We also train with group consistency loss \cite{gu2020pair} for the final epochs. The loss encourages the composition of predicted deformations over image cycles to be identity mappings. Note that NDV and group consistency losses were also used in the winning submission of Learn2Reg 2024 which was also based on the SITReg architecture. The strategies are documented by the GitHub repository \url{https://github.com/honkamj/SITReg}.

The training setup is implemented in PyTorch and we trained the models with A100 and H100 GPUs using Adam as an optimizer. For the group consistency training included for final epochs we used 3 GPUs per training since the loss computation did not easily fit on a single GPU. The earlier epochs we trained on a single GPU.

\section{Results}\label{sec:results}

In Table \ref{table:results} the results of the LUMIR 2025 validation set are shown for the different ablations. The dataset\cite{dufumier2022openbhb,taha2023magnetic} used for training the network consists of T1-weighted brain MRI images. The validation set also consists of brain MRI images, but the out-of-domain set contains T1-weighted images from a different dataset, as well as T1-weighted images with different MRI field strengths. The multi-modal set consists of pairs of T1- and T2-weighted images. Dice overlap and HdDist95 (95\% quantile of Hausdorff distance) are based on segmentations of over $100$ anatomical structures, whereas TRE (target registration error) is based on manual landmarks.

Using MIND features as input representation causes only a very minor drop in in-domain and out-of-domain performance while significantly improving multi-modal performance. The additional strategies systematically improve the performance, although the improvements are not very large. It is noteworthy that the clearly larger non-diffeomorphic volume (NDV) in the baseline version compared to the ones using the MIND feature representation is explained by the multi-modal pairs for which the model predicts very unrealistic deformations.

\begin{table}[t]
\label{table:results}
\caption{Results showcasing the effects of the proposed design choices on the validation set. The values and metrics are directly from the LUMIR 2025 challenge leaderboard (the method holds the 1st place 2 weeks before the challenge test submission is closed). Please refer to Section \ref{sec:results} and the challenge for more details on the metrics. MIND: Transform the input images using the MIND transformation. NDV: Use loss penalizing non-diffeomorphic volume. GC: Use group-consistency loss over image triplets. AUG: Augment input images. ENS: Use ensemble of 5 models.}\label{tab1}
{\fontsize{8}{10}\selectfont
\begin{tabular}{ccccccccccc}
\toprule
\multirow[c]{2}{*}{\rotatebox[origin=c]{90}{MIND}} & \multirow[c]{2}{*}{\rotatebox[origin=c]{90}{NDV}} & \multirow[c]{2}{*}{\rotatebox[origin=c]{90}{GC}} & \multirow[c]{2}{*}{\rotatebox[origin=c]{90}{AUG}} & \multirow[c]{2}{*}{\rotatebox[origin=c]{90}{ENS}} & \multicolumn{3}{c}{Dice$(\%) \uparrow$} & TRE $\downarrow$ & HdDist95 $\downarrow$ & NDV $\downarrow$\\
\cmidrule(r){6-8}\cmidrule(r){9-9}\cmidrule(r){10-10}\cmidrule(r){11-11}
 & & & & & In-domain & Out-of-domain & Multi-modal & In-domain & Overall & Overall\\
\midrule
 \nocheck & \nocheck & \nocheck & \nocheck & \nocheck & 77.7(1.5) & 76.2(1.5) & 28.4(1.5) & 2.30(0.32) & 4.62(1.88) &  	0.052(0.042) \\ % Baseline
 \tablecheck & \nocheck & \nocheck & \nocheck & \nocheck & 77.6(1.4) & 75.9(1.2) & 73.3(2.8) & 2.31(0.30) & 3.18(0.30) & 0.015(0.0022)\\ % a
 \tablecheck & \tablecheck & \tablecheck & \nocheck & \nocheck & 78.0(1.5) & 76.0(1.5) & 73.7(2.8) & 2.27(0.26) & 3.02(0.36) & 0.0017($\expnumber{3.2}{-4}$) \\ % c.0
 \tablecheck & \tablecheck & \tablecheck & \tablecheck & \nocheck & 78.0(1.7) & 76.2(1.1) & 74.2(2.9) & 2.26(0.25) & 2.99(0.33) & 0.0025($\expnumber{4.2}{-4}$) \\ % d.0
 \tablecheck & \tablecheck & \tablecheck & \tablecheck & \tablecheck & \textbf{78.3}(1.7) & \textbf{76.5}(1.2) & \textbf{74.5}(3.0) & \textbf{2.24}(0.27) &  \textbf{2.95}(0.34) & \textbf{0.0015}($\expnumber{3.3}{-4}$) \\ % e
\bottomrule
\end{tabular}
}
\end{table}

\section{Discussion}

The paper proposes a simple deep learning strategy that allows registration of T1 and T2 weighted MRI scans while training only on T1-weighted MRI scans by transforming the inputs with the MIND transformation before feeding them into the network. Good results indicate that the MIND transformation transforms T1- and T2-weighted MRI images into relatively similar representations. The performance of T1-T2 registration with the proposed method, while close, is still worse than the in-domain performance. A potential future research direction is hence to look for even more suitable input transformations. Further research is also needed on the performance of the method on other modalities or anatomies.

\begin{credits}
\subsubsection{\ackname} This work was supported by the Research Council of Finland (Flagship programme: Finnish Center for Artificial Intelligence FCAI, and grants 352986, 358246) and EU (H2020 grant 101016775 and NextGenerationEU). We also acknowledge the computational resources provided by the Aalto Science-IT Project.

\subsubsection{\discintname}
The authors have no competing interests to declare that are relevant to the content of this article.
\end{credits}

\bibliographystyle{splncs04}
\bibliography{main}

\begin{thebibliography}{10}
\providecommand{\url}[1]{\texttt{#1}}
\providecommand{\urlprefix}{URL }
\providecommand{\doi}[1]{https://doi.org/#1}

\bibitem{chen2022transmorph}
Chen, J., Frey, E.C., He, Y., Segars, W.P., Li, Y., Du, Y.: Transmorph: Transformer for unsupervised medical image registration. Medical image analysis  \textbf{82},  102615 (2022)

\bibitem{dufumier2022openbhb}
Dufumier, B., Grigis, A., Victor, J., Ambroise, C., Frouin, V., Duchesnay, E.: Openbhb: a large-scale multi-site brain mri data-set for age prediction and debiasing. NeuroImage  \textbf{263},  119637 (2022)

\bibitem{gu2020pair}
Gu, D., Cao, X., Ma, S., Chen, L., Liu, G., Shen, D., Xue, Z.: Pair-wise and group-wise deformation consistency in deep registration network. In: International Conference on Medical Image Computing and Computer-Assisted Intervention. pp. 171--180. Springer (2020)

\bibitem{guo2019multi}
Guo, C.K.: Multi-modal image registration with unsupervised deep learning. Ph.D. thesis, Massachusetts Institute of Technology (2019)

\bibitem{heinrich2012mind}
Heinrich, M.P., Jenkinson, M., Bhushan, M., Matin, T., Gleeson, F.V., Brady, M., Schnabel, J.A.: Mind: Modality independent neighbourhood descriptor for multi-modal deformable registration. Medical image analysis  \textbf{16}(7),  1423--1435 (2012)

\bibitem{hering2022learn2reg}
Hering, A., Hansen, L., Mok, T.C., Chung, A.C., Siebert, H., H{\"a}ger, S., Lange, A., Kuckertz, S., Heldmann, S., Shao, W., et~al.: Learn2reg: comprehensive multi-task medical image registration challenge, dataset and evaluation in the era of deep learning. IEEE Transactions on Medical Imaging  \textbf{42}(3),  697--712 (2022)

\bibitem{honkamaa2023sitreg}
Honkamaa, J., Marttinen, P.: Sitreg: Multi-resolution architecture for symmetric, inverse consistent, and topology preserving image registration. arXiv preprint arXiv:2303.10211  (2023)

\bibitem{jena2024deep}
Jena, R., Sethi, D., Chaudhari, P., Gee, J.: Deep learning in medical image registration: Magic or mirage? Advances in Neural Information Processing Systems  \textbf{37},  108331--108353 (2024)

\bibitem{liu2024finite}
Liu, Y., Chen, J., Wei, S., Carass, A., Prince, J.: On finite difference jacobian computation in deformable image registration. International journal of computer vision  \textbf{132}(9),  3678--3688 (2024)

\bibitem{taha2023magnetic}
Taha, A., Gilmore, G., Abbass, M., Kai, J., Kuehn, T., Demarco, J., Gupta, G., Zajner, C., Cao, D., Chevalier, R., et~al.: Magnetic resonance imaging datasets with anatomical fiducials for quality control and registration. Scientific Data  \textbf{10}(1), ~449 (2023)

\end{thebibliography}
\end{document}